\documentclass[runningheads]{llncs}
\usepackage[T1]{fontenc}
\usepackage{graphicx}
\usepackage[acronym, symbols]{glossaries}
\usepackage{hyperref}
\usepackage{subcaption}
\usepackage{array}
\newcolumntype{P}[1]{>{\centering\arraybackslash}p{#1}}
\usepackage{booktabs}
\usepackage[justification=centering]{caption}
\usepackage{listings}
\usepackage{amsmath}
\usepackage{amssymb}
\usepackage{algorithm}
\usepackage{algpseudocode}
\usepackage{tablefootnote}
\usepackage{xcolor}
\usepackage{listings}
\usepackage{dcolumn}
\usepackage{siunitx}
\definecolor{codegreen}{rgb}{0,0.6,0}
\makeglossaries
\newacronym{kg}{KG}{Knowledge Graph}
\newacronym{bfs}{BFS}{Breadth First Search}
\newacronym{dfs}{DFS}{Depth First Search}
\newcolumntype{d}[1]{D{.}{.}{#1}}
\begin{document}
\title{gpuRDF2vec -- Scalable GPU-based RDF2vec}
%
%
\author{Martin B\"ockling\inst{1}\orcidID{0000-0002-1143-4686} \and
Heiko Paulheim\inst{1}\orcidID{0000-0003-4386-8195}}
%
%
%
\institute{Data and Web Science Group, University of Mannheim, Mannheim 68160, Germany}
\maketitle              
\begin{abstract}
Generating \gls*{kg} embeddings at web scale remains challenging. Among existing techniques, RDF2vec combines effectiveness with strong scalability. We present gpuRDF2vec, an open source library that harnesses modern GPUs and supports multi‑node execution to accelerate every stage of the RDF2vec pipeline. Extensive experiments on both synthetically generated graphs and real‑world benchmarks show that gpuRDF2vec achieves up to a substantial speedup over the currently fastest alternative, i.e., jRDF2vec. In a single‑node setup, our walk‑extraction phase alone outperforms pyRDF2vec, SparkKGML, and jRDF2vec by a substantial margin using random walks on large/ dense graphs, and scales very well to longer walks, which typically lead to better quality embeddings. Our implementation of gpuRDF2vec enables practitioners and researchers to train high‑quality KG embeddings on large‑scale graphs within practical time budgets and builds on top of Pytorch Lightning for the scalable word2vec implementation \footnote{Our github repository can be found under the \href{https://doi.org/10.5281/zenodo.15700011}{following link} and can be downloaded as a \href{https://pypi.org/project/rdf2vecgpu/}{pypi package}.}.

\end{abstract}

\section{Introduction}
Knowledge Graphs are a key component of many applications that require access to large amounts of domain-independent or domain-specific information~\cite{noy2019industry}. Open knowledge graphs, such as DBpedia~\cite{auer2007dbpedia} and Wikidata~\cite{vrandevcic2014wikidata}, as well as domain-specific sources like the Gene Ontology~\cite{gene2019gene}, serve as valuable resources for building knowledge-intensive applications.

Many downstream applications rely on knowledge graph embeddings---dense numeric vector representations of entities in knowledge graphs---to make predictions, compute similarities between entities, and support search and retrieval~\cite{portisch2022knowledge}. While such embeddings are valuable, computing them at scale remains challenging~\cite{biswas2023knowledge}. For example, in the case of Wikidata, to the best of our knowledge, there are no pre-trained embeddings available for versions newer than five years old.

In this paper, we address the scalability challenges of knowledge graph embedding methods. We propose an implementation of the widely used RDF2vec method for knowledge graph embedding~\cite{paulheim2023embedding} that can run on GPUs, thereby enabling significant performance gains. While already outperforming existing implementations on a single GPU node, our approach can also leverage multiple nodes for additional speed-ups. We demonstrate that with \textit{gpuRDF2vec}, a complete RDF2vec embedding of a recent version of Wikidata can be computed in under 24 hours.

\section{Background and Related Resources}

RDF2vec is a \gls*{kg} embedding model composed of two main steps. First, graph patterns are extracted from the \gls*{kg} to generate a corpus consisting of graph walks. In the original implementation, two different walk strategies were used. The first strategy involves random walks, while the second leverages Weisfeiler-Lehman subtree kernels to produce walks from various nodes~\cite{Ristoski2016}. Later implementations added a third strategy, i.e., graph walks generated by \gls*{bfs} or \gls*{dfs}, which enumerate all possible walks from each node, and then optionally sample from those to restrict the overall number of walks. In the second step of RDF2vec, the Word2Vec model is employed to the extracted walks to generate vector representations for individual nodes as well as edges. RDF2vec has been applied to a variety of \gls*{kg}-related tasks, such as link prediction and entity classification. While RDF2vec has demonstrated competitive performance across different benchmarks, it has also shown good scalability, particularly on larger graphs. Over time, several implementations of RDF2vec have been developed to extend and adapt the original release. 

PyRDF2vec is a purely Python-based library that uses a modular structure to enable easy adaptability. It implements various sampling strategies and multiple walk algorithms to facilitate path extraction.~\cite{steenwinckel2021walk} Ultimately, the extracted walks are used in conjunction with a Word2Vec model provided by Gensim~\cite{Steenwinckel2023}. Upon inspecting the published source code, it becomes evident that the authors have implemented both \gls*{bfs} and \gls*{dfs} variations for walk extraction. PyRDF2vec samples a fixed number of walks from the resulting \gls*{bfs} or \gls*{dfs} walks, introducing randomness only when a node has more potential walks than the allowed limit. Since random walks in the literature are typically understood as an iterative random selection of the next hop, we refer to PyRDF2vec's approach as a \gls*{bfs}- or \gls*{dfs}-based implementation in this context. Additionally, pyRDF2vec supports random walks weighted by graph metrics such as graph centrality~\cite{cochez2017biased}.

jRDF2vec is a Java-based implementation of RDF2vec. It implements walk strategies in Java, while the Word2Vec component is handled either by Gensim or a C++-based Word2Vec implementation to support an order-aware Word2Vec model~\cite{Ling2015,Portisch2021}. jRDF2vec includes four different walk strategies, allowing adaptation of the graph path extraction process~\cite{portisch2022walk}, along with variants that ensure duplicate-free random walks~\cite{Portisch2021}. Overall, jRDF2vec has demonstrated solid runtime performance and is capable of scaling to \glspl*{kg} such as DBpedia~\cite{Portisch2021,portisch2022,egami.2023}.

A recent implementation of RDF2vec leverages Spark to distribute both the walk extraction and the Word2Vec training. The library \textit{sparkkgml} builds on top of GraphFrames, a Spark-based graph processing component. To extract walks, \textit{sparkkgml} uses motif walks to identify paths within the \gls*{kg}. These motif walks begin from a given node and expand based on a specified walking depth. As the depth increases, the motif string pattern is extended to expand the search space for walk extraction. In general, \textit{sparkkgml} implements \gls*{bfs} walk extraction along with variations such as entity and property walks~
\cite{portisch2022knowledge}. For vectorization, the generated path sequences are stored in a Spark DataFrame, which is then processed using the Word2Vec implementation from Spark MLlib~\cite{Gergin2024}.

A detailed overview of the various capabilities across all RDF2vec libraries is provided in \autoref{tab:library_comparison}.

\begin{table}[ht]
    \centering
    \caption{Comparison of capabilities within the different RDF2vec packages}
    \begin{tabular}{lcccc}
        \toprule
        \textbf{Capabilities} & \textbf{jRDF2vec} & \textbf{pyRDF2vec} & \textbf{sparkkgml} & \textbf{gpuRDF2vec} \\
        \midrule
        Random Walk & \checkmark & $\times$ & $\times$& \checkmark \\
        Random Walk duplicate free & \checkmark & $\times$ & $\times$ & \checkmark \\
        Centrality ranked walks & $\times$ & \checkmark & $\times$ & $\times$ \\
        Property Walks & \checkmark & $\times$ & \checkmark & \checkmark \\
        Entity Walks & \checkmark & $\times$ & \checkmark & \checkmark \\
        BFS  & $\times$ & \checkmark & \checkmark & \checkmark \\
        DFS  & $\times$ & \checkmark & $\times$& $\times$\\
        Multi-node Scalability  & $\times$ & $\times$ & \checkmark & \checkmark \\
    \end{tabular}
    \label{tab:library_comparison}
\end{table}


\section{The gpuRDF2vec Package} \label{cha:ResourceOverview}
We provide a package that supports both walk generation and vectorization methods for the extracted graph walks. Our graph-related components are built on top of cuGraph~\cite{fender2022rapids}, while data transformations within the package utilize cuDF~\cite{rapids.2023}. Multiple benchmarks have demonstrated that both libraries outperform their respective counterparts in terms of execution speed~\cite{Hernandez2020,rapids.2023a,rapids.2023}. 

\subsection{Walk Extraction} \label{cha:WalkExtraction}
For extracting individual paths from the provided \gls*{kg}, we use cuGraph as the computational backbone. With the exception of the \textit{.nt} file format, we currently do not provide direct RDF parsing integration with the RAPIDS libraries, but rely on rdflib. This first loads the \gls*{kg} into device memory and then loads it into the GPU. In addition to traditional 
\gls*{kg} file formats, we support \textit{CSV}, \textit{Text}, \textit{Parquet}, and \textit{ORC} file formats. All supported read operations are handled by the cuDF library, which can be scaled using Dask over multi-node GPU clusters ~\cite{rocklin2015dask}.

\begin{figure}[t]
    \centering
    \includegraphics[width=0.95\linewidth]{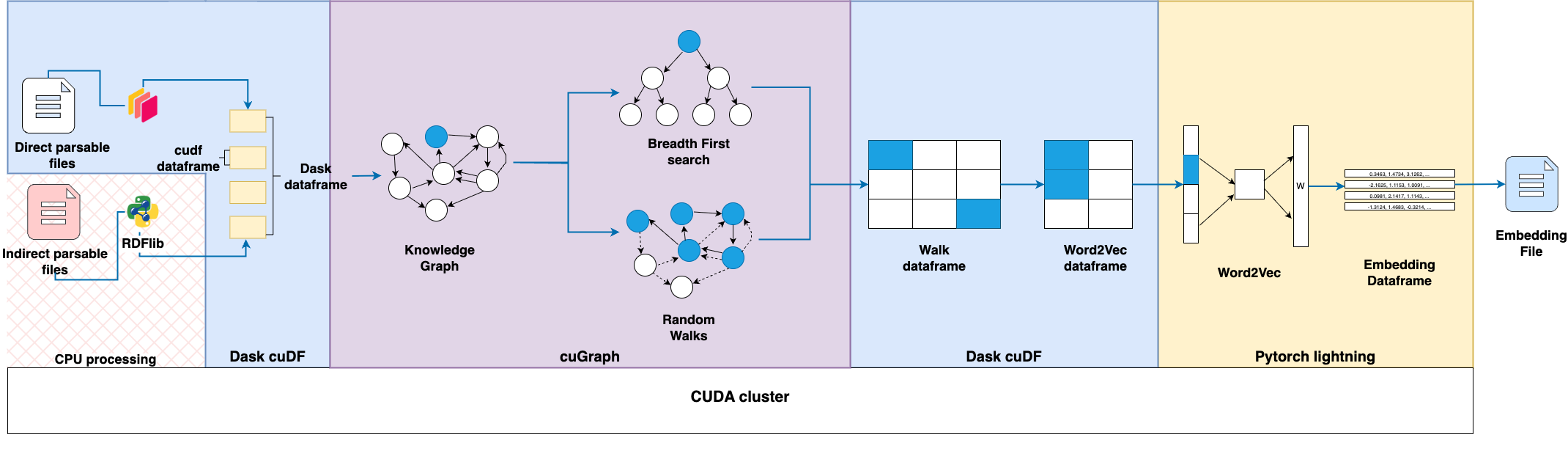}
    \caption{Schema diagram of gpuRDF2Vec approach using GPU clusters for computing RDF2Vec embeddings}
    \label{fig:rdf2vecApproach}
\end{figure}

cuGraph supports running a selected set of algorithms in a distributed GPU setup by leveraging Dask as the distribution framework. Both random walks and \gls*{bfs} can be executed over Dask to enable multi-node scalability for walk extraction. At the very beginning we build up our vocabulary and each individual entity and predicate is tokenized at the beginning of our pipeline.

Both the random walk and \gls*{bfs} traversal implementations in cuGraph are built for unlabeled graphs and thus offer no direct support for \glspl*{kg}, i.e., they only extract sequences of entities during traversal. In our environment, we have adapted these algorithms to better suit knowledge graph traversal. In the following paragraphs, we present an overview of our implementation, in which we leverage CUDA kernels provided by both cuDF and cuGraph for all computations.

For the random walk implementation, we receive a sequence of nodes where each walk has a fixed length. If a random walk terminates before reaching the defined maximum depth, it is padded with \texttt{-1}. This necessitates custom transformation logic to align the generated node sequences with the desired triple-based extraction. Once the random walk results are retrieved, each padding within the walk is removed for further processing.

For \gls*{bfs}, we start with the predecessor vector returned by \textit{cuGraph}'s \gls*{bfs} method. We then convert the rooted tree into an \emph{edge list duplicated per walk}, so that every path from a leaf to the root is explicitly materialized. \textit{cuGraph} stores a unique predecessor $p(v)$ for each vertex $v$ --- i.e., the first vertex to reach $v$ in the \gls*{bfs} frontier. This deterministic behavior ensures that each vertex appears in one walk in maximum once. All rows in the resulting \gls*{bfs} dataframe that are not root nodes are projected into source-target relationships. To reconstruct full walk paths, we identify the leaf nodes in the \gls*{bfs} spanning tree and iteratively join the current set of frontier nodes with their predecessors. 

The resulting table \textit{(source, target, walk\_id)} contains each edge once for every leaf-to-root path, reproducing the traversal order required for visualizing end-to-end retrieval chains while preserving the single-parent semantics of the original \gls*{bfs} extraction.

The resulting dataframes from both the random walk and the \gls*{bfs} traversal serve as the basis for the data loader used in our Word2Vec implementation. 

\subsection{Word2Vec implementation} \label{cha:Word2VecImpl}
The library includes PyTorch Lightning implementations for both Skip-Gram and Continuous Bag of Words (CBOW) that run natively on CUDA.  Our \emph{Skip-Gram with Negative Sampling} (SGNS) implementation uses a compact skipgram data module, which converts the walk corpus into center–context pairs and automatically tunes the mini-batch size (default: $1/20$ of the corpus), while supporting multi-worker prefetching and page-pinned memory transfer to the GPU. 

The core model, \texttt{SkipGram}, maintains separate input and output embedding tables, both initialized from a uniform distribution $\mathcal{U}(-1/d, 1/d)$ with embedding dimension~$d$, and stored as \texttt{nn.Embedding(\dots, sparse{=}True)}. This sparse CSR backend ensures that only the rows sampled in the current batch are updated, reducing gradient memory traffic by up to $90\%$ compared to dense \texttt{Adam}. During each \texttt{training\_step}, the module draws $k$ i.i.d.\ negative samples from a uniform alias table, computes the positive and negative logits, and minimizes the binary cross-entropy loss in a single fused kernel. Gradients are propagated using \texttt{SparseAdam}, unless the user explicitly opts for dense \texttt{Adam}.

Since the class inherits from \texttt{LightningModule}, users can easily swap optimizers, learning rate schedulers, or precision plugins (e.g., mixed FP16) without modifying the underlying implementation. The entire SGNS implementations integrates seamlessly into the broader pipeline via the generic \texttt{fit\_transform} interface described earlier.

Our \emph{Continuous Bag-of-Words} (CBOW) implementation mirrors the design philosophy of the Skip-Gram model, but reverses the prediction direction: it learns to predict the \emph{target} (center) word from its surrounding context. It maintains two independent embedding matrices: \texttt{in\_embeddings} for context words and \texttt{output\_embeddings} for target words, both initialized uniformly in the interval \(\bigl[-\tfrac{1}{d}, +\tfrac{1}{d}\bigr]\), where \(d\) is the embedding dimension. This ensures a small yet symmetric initialization range that helps accelerate early convergence.

During the forward pass, we first obtain dense context and center representations, then compute positive logits via the dot product \(\langle\mathbf{c},\mathbf{t}\rangle\), while negative logits are formed by contrasting the same center vector against \(k=\textit{window\_size}\) noise samples drawn uniformly from the vocabulary (negative sampling).  
A binary cross-entropy loss with logits is applied independently to positive and negative pairs and averaged across the mini-batch:
\[
\mathcal{L}
  = \frac{1}{B}\sum_{i=1}^{B}
      \bigl(
        \mathrm{BCE}( \langle\mathbf{c}_i,\mathbf{t}_i\rangle, 1)
        + \mathrm{BCE}\bigl( \mathbf{c}_i^{\!\top}\mathbf{n}_i, 0\bigr)
      \bigr),
\]
where \(B\) is the batch size and \(\mathbf{n}_i\) stacks the \(k\) negative samples for instance \(i\).  

Training is orchestrated through PyTorch Lightning’s \texttt{training\_step}, which dynamically generates fresh negative samples on-the-fly for each batch, promoting robustness and memory efficiency.  
Like for SGNS, the optimisation method defaults to \texttt{SparseAdam} capitalising on the natural sparsity of word-embedding gradients—yet falls back to standard \texttt{Adam} when the \texttt{use\_sparse} flag is disabled, both driven by a user-defined learning rate.  
Overall, this concise, GPU-friendly module combines efficient sampling, sparse optimisation support and Lightning’s high-level abstractions, making it straightforward to scale to very large vocabularies.

\subsection{Usage Example}

gpuRDF2vec exposes every major decision point of the pipeline as a constructor argument, so users can tailor both \textit{walk generation} and \textit{embedding training} without having to alter the core code.
The walker can be switched from the default \texttt{walk\_strategy=,"random"} to breadth‑first, weighted, or any user‑defined strategy simply by passing the fully‑qualified class name; depth (\texttt{walk\_depth}), fan‑out (\texttt{walk\_number}), and reproducibility (\texttt{random\_state}, \texttt{reproducible}) offer fine‑grained control over corpus creation.
On the embedding side, the model backend is pluggable via \texttt{embedding\_model} (currently \texttt{"skipgram"} or \texttt{"cbow"}, and all Word2Vec hyperparameters map one‑to‑one to their PyTorch implementations, ensuring easy transfer of tuned values from earlier studies.
Performance knobs include adaptive \texttt{batch\_size} (auto‑scales to GPU RAM per default), optional multi‑GPU training (\texttt{multi\_gpu=true}, NCCL backend), and CPU fallback (\texttt{cpu\_count}) for feature‑engineering steps; any out‑of‑memory event triggers an automatic micro‑batch fallback rather than aborting the run.

Extensibility follows the \emph{open‑module} pattern: new walkers or samplers only need to subclass \texttt{BaseWalker} and register themselves once, after which they can be referenced by name in the constructor; alternative optimisers or loss functions can be injected through the \texttt{optimizer\_cls} and \texttt{loss\_fn} callables exposed in \texttt{fit\_transform}.
Post‑training artefacts—embeddings, training logs, TensorBoard traces—are emitted conditionally via \texttt{generate\_artifact}, and the save‑hook API lets downstream pipelines intercept vectors as NumPy arrays, Pandas DataFrames, or Arrow tables.
All entry points are documented with type hints, unit tests, and minimal examples, so integrating custom logic (e.g., path‑context filters or graph‑specific negative sampling) typically requires fewer than 50 lines of new code. An example for the constructor call of our class can be found in \autoref{lst:gpu_rdf2vec_example}.

\begin{lstlisting}[language=Python, caption=Python library minimal example,label={lst:gpu_rdf2vec_example}]
from rdf2vecgpu.gpu_rdf2vec import GPU_RDF2Vec

# Initialize model
model = GPU_RDF2vec(
    walk_strategy="random",
    walk_depth=5,
    walk_number=100,
    embedding_model="skipgram",
    epochs=5,
    batch_size=None,
    vector_size=100,
    window_size=5,
    min_count=10,
    learning_rate=0.01,
    negative_samples=5,
    random_state=42,
    reproducible=False,
    multi_gpu=False,
    generate_artifact=False,
    cpu_count=20
)

# Load edge data from 'path'
edge_data = model.load_data(path)

# Fit and transform to get embeddings
embeddings = model.fit_transform(
    edge_df=edge_data,
    walk_vertices=None
)
\end{lstlisting}


\section{Implementation Details and Optimization} \label{cha:TechnicalDetails}
There are a few engineering choices that make \textsc{gpuRDF2vec} reach order--of--magnitude speed‑ups over CPU‑bound implementations while keeping memory consumption predictable, both related to the walk extraction as well as to the training of the Word2vec model.


\subsection{Optimizing Walk Extraction} \label{cha:WalkExtractionOptimization}
For both random walks as well as \gls*{bfs} walks, the majority of the operations happen over the cuDF and cuGraph functions. The primary motivation for executing operations on GPU kernels and minimizing the data transfer is to maintain optimal performance during the walk extraction process as well as the path construction \cite{cudfPerformance}. By minimizing the interactions and data transfers between the GPU and CPU, we prevent potential bottlenecks and latency ensuring that the path construction phase proceeds without unnecessary overhead.

To efficiently generate multiple walks per entity, we avoid iterating repeatedly through a list of entities. Instead, we replicate the indices within a cuDF DataFrame with the number of walks. This replication strategy leverages GPU parallelism by enabling simultaneous processing of multiple walks per entity. This significantly improves the runtime performance of random walk generation over the complete graph. By fully utilizing the parallel capabilities of GPU hardware, we ensure that the overhead typically associated with iterative Python-based approaches is eliminated. In an overall comparison, we have seen that the speedup between the iteration based approach and the iteration-based approach is nearly 15 times faster on our infrastructure.

In the context of \gls*{bfs}, the traversal of walks currently involves iterative recursive joins. Although these join operations are executed on GPU kernels, our current implementation introduces some overhead associated with constructing the walks from \gls*{bfs} traversals. Future enhancements will aim to fully exploit CUDA capabilities to reconstruct walks directly, thus eliminating the need for recursive joins, thereby further optimizing performance.

\subsection{Handing From cuDF to Pytorch} \label{cha:cudfToPytorch}
To fully exploit our GPU-centric architecture, we redesigned the standard PyTorch Lightning data loading pipeline. In particular, we evaluated the built-in CPU-parallelization capabilities and identified several performance bottlenecks and opportunities for improvement.

First, PyTorch Lightning’s native approach parallelizes data loader operations across multiple CPU cores. While this is beneficial for the typical scenario of loading data from disk, it introduces unnecessary overhead on our GPU-resident datasets: each batch is first assembled in main memory and then transferred to the device, delaying the start of GPU computations. In our experiments, this ``two-stage'' handoff lengthened overall epoch time due to redundant memory copies and synchronization delays.

To alleviate this, we replaced the default CPU‐based loading with a cuDF‐backed approach. We store the \textit{context} and \textit{center} columns in a single cuDF DataFrame on the GPU, and export them as independent DLPack tensors. Initially, each batch construction performs a deep copy into these tensors, temporarily increasing VRAM usage. However, once resident on the GPU, subsequent sampling and preprocessing steps can be executed entirely on the device, eliminating PCIe transfer latency and letting CUDA kernels manage parallelization across SMs.

Building on this, we further optimized the DataLoader to dispense only tensor indices rather than full slices. At load time, each worker simply retrieves an index, and the GPU tensor is directly indexed in place. This ``index‐only'' strategy incurs virtually no copy overhead, since CUDA’s memory model supports constant‐time pointer arithmetic. Although we disable CPU‐side parallelism in favor of streamlined GPU operations, this change reduced the DataLoader’s share of total training time from approximately 85\% down to near parity with the model’s forward/backward passes. The result is a balanced pipeline in which data provisioning and model computation proceed concurrently with minimal idle periods on either side.

\subsection{Optimizing Word2Vec Training} \label{cha:pytorch}
To streamline memory provisioning and reduce tuning iterations, we employ a power‐guessing heuristic for initial batch size selection. Specifically, we estimate the per‐sample GPU footprint via our cuDF loader, then divide the total available VRAM by four times this footprint as a first‐pass batch size. This “divide‐by‐four” rule of thumb drastically narrows the search space for subsequent fine‐grained batch size optimization, cutting down on wasted training runs and manual tuning.

Next, we migrated all sampling and preprocessing kernels into native PyTorch tensor operations, removing Python‐level loops and function call overhead from the critical execution path. By leveraging PyTorch’s C++ back end for indexing, masking, and tensor transformations, we eliminate the Global Interpreter Lock (GIL) bottleneck and reduce CPU utilization, resulting in more consistent throughput and lower end‐to‐end latency.

Finally, to avoid out‐of‐memory (OOM) failures during training, we integrate a dynamic safeguard that continuously monitors estimated VRAM usage and enforces a hard cap at 90\% of total GPU memory. If the predicted utilization exceeds this threshold, the loader automatically scales back the batch size or pauses new allocations, ensuring stable training without manual intervention. These combined measures yield a robust, low‐overhead pipeline that adapts to different hardware profiles while minimizing memory‐related disruptions.  

We adopt a data‑parallel strategy implemented with PyTorch Distributed + NCCL: each GPU maintains an identical KG shard but a disjoint walk corpus.  Gradients are all‑reduced every \texttt{sync\_interval} (${\sim}500$ ms by default) to amortise PCIe/NVLink traffic. For clusters, the number of nodes in the training loop are defined to distribute the path.


\section{Experimental Setup}

For our experiments, we utilize a dedicated virtual machine equipped with a single NVIDIA A100 GPU and 20 vCPUs based on Intel(R) Xeon(R) Silver 4114 CPUs running at 2.20\,GHz. Each experimental run has access to a total of 200\,GB of RAM. The VM runs the AlmaLinux 9.5 distribution. 

Since the different RDF2vec implementations we compare have varying software requirements, we provide a comprehensive list of dependencies for each implementation in the following \href{https://github.com/MartinBoeckling/rdf2vecgpu/tree/performance_dev/performance/env_files}{GitHub folder}.

\subsection{Knowledge Graphs}
We compare two categories of \glspl*{kg} in our setup: synthetically generated graphs and a real-world benchmark graph.

For the synthetic graphs, we use the \textit{igraph} library due to its low memory footprint. As \textit{igraph} does not natively support edge labels, we randomly assign each edge a predicate from a fixed set~\cite{Csrdi2006TheIS}. Three generation models are used:

\textbf{Barabási-Albert (BA):} This model captures heavy-tailed degree distributions common in real-world graphs~\cite{Barabasi1999}. Starting from a fully connected seed graph, new vertices are iteratively added, each connecting to existing nodes with a probability proportional to their degree. We generate three BA graphs using a connection probability of 0.4.

\textbf{Erd\H{o}s-Rényi (ER):} In this model~\cite{Erdoes1959}, edges are formed independently with a uniform probability $p$ between any pair of $n$ nodes. Degree distributions follow a binomial (or Poisson in the limit), resulting in narrow degree variance and high clustering.

\textbf{Uniform-Attachment (UA):} This growing graph model is a dynamic variant of ER, where new vertices connect to existing ones uniformly at random, regardless of degree~\cite{Csrdi2006TheIS}. Depending on configuration (e.g., the \texttt{citation} flag in \texttt{igraph::sample\_growing()}), the structure varies between recursive trees and looser networks. The resulting degree distribution is exponential, and typical distances scale as $\mathcal{O}(\log n)$.

The set of all synthetic graphs used for evaluation can be found under the following ~\cite{Boeckling2025}.

\textbf{FB15k-237:} A widely used benchmark derived from Freebase~\cite{Toutanova2015}, FB15k-237 contains 14,951 entities and 1,345 relation types across 592,213 triples. These are split into 483,142 training, 50,000 validation, and 59,071 test facts. The dataset spans multiple domains, including people, locations, music, and film, with both one-to-one and many-to-many relations.

\textbf{Wikidata5M} is a large‐scale knowledge-graph corpus distilled from the public Wikidata 2021 dump and aligned with Wikipedia text. It contains \num{4.59} M entities and \num{822} relation types, yielding \num{20.6} M training triples plus \num{5\,163}/\num{5\,133} triples for validation and test, respectively, in the standard transductive split, with analogous inductive splits for open-world evaluation~\cite{Wang2019}. Each entity is paired with the full text of its Wikipedia article, enabling joint modeling of structured triples and unstructured descriptions and making the dataset a strong benchmark for GPU-accelerated RDF2Vec embeddings.

A detailed comparison of the characteristics of all KGs is provided in \autoref{tab:KGComparison}. The calculation of the average betweenness centrality for wikidata5m timed out after 4 hours running on a Nvidia H100 GPU. We therefore mark the overall record as

\begin{table}[t]
    \centering
    \caption{Statistics of benchmarked Knowledge Graphs}
    \label{tab:KGComparison}
    \begin{tabular}{lrrd{4.4}d{5.4}d{1.4}}
        \toprule
        \textbf{KG} & \textbf{Vertices} & \textbf{Edges} & \multicolumn{1}{c}{\textbf{Avg. degree}} & \multicolumn{1}{c}{\textbf{Avg. Betweenness}} & \multicolumn{1}{c}{\textbf{Density}}\\
        \midrule
        Barabasi 100 & 100 & 99 & 1.98 & 3.64 & 0.01\\
        Barabasi 1000 & 1,000 & 999 & 1.998 & 6.506 & 0.001\\
        Barabasi 10000 & 10,000 & 9,999 & 1.9998 & 11.5997 & 0.0001\\
        Erdos Renyi 100 & 100 & 3,887 & 77.74 & 60.13 & 0.3926 \\
        Erdos Renyi 1000 & 1,000 & 399,727 & 799.454 & 599.273 & 0.4001 \\
        Erdos Renyi 10000 & 10,000 & 39,992,483 & 7998.4966 & 5999.7517 & 0.3999 \\
        Random 100 & 93 & 990 & 15.0538 & 33.3871 & 0.1157\\
        Random 1000 & 951 & 9,990 & 19.6172 & 572.0578 & 0.0111\\
        Random 10000 & 9,541 & 99,990 & 20.7259 & 7310.749 & 0.0011\\
        FB-15k 237 & 14,541 & 310,116 & 39.0438 & 32486.8515 & 0.0015\\
        Wikidata-5m & 4,594,485 & 20,624,575 & 8.9780 & \varnothing & 0.0000009\\
        \bottomrule
    \end{tabular}
\end{table}

\subsection{Experimental Settings}

If an individual run of any of the RDF2vec libraries exceeds 4 hours of processing time, we consider the RDF2vec run as not successful. 
In total, we run ten RDF2vec embedding generations independently with randomly set seeds in order to also analyze stability.

For the overall experimental setup, we run the overall experiment on different environments due to the conflicting dependencies between the different RDF2vec implementations. A detailed overview of the different environment settings can be found under the following \href{https://github.com/MartinBoeckling/rdf2vecgpu/tree/main/performance/env_files}{Github folder}. For pyRDF2Vec, we used the latest version with 0.2.3 which supports as a maximum major Python version 3.10. For jRDF2Vec, we used the newest compiled jar file (version of 1.3) and a subsequent Python version of 3.12. For sparkkgml in version 0.2.2, the major Python version is 3.11. with a pyspark version 3.5.5. The graphframes which is used for the graph version is set to the newest available JAR spark library which is 0.8.4. For gpuRDF2Vec, we use Python version 3.12 as a base, together with cudf 25.4, cugraph 25.4.1 and Pytorch 2.5.1.


\section{Results and Discussion} \label{sec:results}
For the overall run of the different results, we provide the results on one specific hyperparameter set. We provide the average runtime of ten independent runs together with the overall standard deviation. For the rank column in \autoref{tab:comparison} we assign per dataset for the lowest average runtime the rank 1. The hyperparameters are fixed on a maximum walk depth of 8, a walk number of 500, training the SkipGram model for 10 epochs. A wider set of the hyperparameter sweep can be found under the following \href{https://github.com/MartinBoeckling/rdf2vecgpu/tree/main/performance}{Github page}. To comply with the single GPU setup, we reduced for the evaluation on Wikidata-5m benchmark the hyperparameters to a maximum walk depth of 8, a walk number of 100 and the training of the SkipGram model is going over 10 epochs.

\begin{table}[p]
    \centering
    \caption{Comparison of Different RDF2vec Runs. $\triangle$ denotes a timeout.}
    \label{tab:comparison}
    \begin{tabular}{llcccc}
        \toprule
        \textbf{Package} & \textbf{Dataset} & \textbf{Hardware} & \textbf{Time (s)} $\downarrow$ & \textbf{Rank}\\
        \midrule
        \textbf{pyRDF2vec} & Barabasi 100 & CPU & $\mathbf{2.76\pm1.47}$ & 1 \\
         & Barabasi 1000 & CPU & $\mathbf{3.12\pm1.4}$ & 1 \\
         & Barabasi 10000 & CPU & $\underline{19.53\pm6.37}$ & 2 \\
         & Erdos Renyi 100 & CPU & $\mathbf{25.11\pm5.77}$ & 1 \\
         & Erdos Renyi 1000 & CPU & $1439.02\pm176.88$ & 4 \\
         & Erdos Renyi 10000 & CPU & $\triangle$ & $\varnothing$ \\
         & Random 100 & CPU & $17.43\pm4.65$ & 3 \\
         & Random 1000 & CPU & $142.59\pm43.4$ & 4 \\
         & Random 10000 & CPU & $1476.91\pm337.77$ & 4 \\
         & FB15k & CPU & $8085.93\pm3705.38$ & $3$ \\
         & Wikidata-5m & GPU & $\triangle$ & $\varnothing$ \\
        \textbf{sparkkgml} & Barabasi 100 & CPU & $68.75\pm7.76$ & 4 \\
         & Barabasi 1000 & CPU & $62.49\pm14.92$ & 3 \\
         & Barabasi 10000 & CPU & $209.07\pm45.78$ & 3 \\
         & Erdos Renyi 100 & CPU & $\triangle$ & $\varnothing$ \\
         & Erdos Renyi 1000 & CPU & $\triangle$ & $\varnothing$ \\
         & Erdos Renyi 10000 & CPU & $\triangle$ & $\varnothing$ \\
         & Random 100 & CPU & $\triangle$ & $\varnothing$ \\
         & Random 1000 & CPU & $\triangle$ & $\varnothing$ \\
         & Random 10000 & CPU & $\triangle$ & $\varnothing$ \\
         & FB15k & CPU & $\triangle$ & $\varnothing$ \\
         & Wikidata-5m & GPU & $\triangle$ & $\varnothing$ \\
        \textbf{jRDF2vec} & Barabasi 100 & CPU & $\underline{12.2\pm2.61}$ & 2 \\
         & Barabasi 1000 & CPU & $\underline{11.8\pm0.08}$ & 2 \\
         & Barabasi 10000 & CPU & $\mathbf{12.42\pm0.06}$ & 1 \\
         & Erdos Renyi 100 & CPU & $162\pm744.53$ & 3 \\
         & Erdos Renyi 1000 & CPU & $\underline{522.15\pm8.47}$ & 2 \\
         & Erdos Renyi 10000 & CPU & $\triangle$ & $\varnothing$ \\
         & Random 100 & CPU & $\mathbf{11.89\pm0.07}$ & $1$ \\
         & Random 1000 & CPU & $\mathbf{13.23\pm0.14}$ & $1$ \\
         & Random 10000 & CPU & $\mathbf{21.52 \pm 0.19}$ & $1$ \\
         & FB15k & CPU &  $\mathbf{194.35\pm20.06}$ & $1$ \\
         & Wikidata-5m & GPU & $\triangle$ & $\varnothing$ \\
        \textbf{gpuRDF2vec} & Barabasi 100 & GPU & $25\pm8.01$ & 3 \\
         & Barabasi 1000 & GPU & $159.36\pm56.54$ & 4 \\
         & Barabasi 10000 & GPU & $1964.49\pm294.48$ & 4 \\
         & Erdos Renyi 100 & GPU & $\underline{32.43\pm5.93}$ & 2 \\
         & Erdos Renyi 1000 & GPU & $\mathbf{292.92 \pm13.18}$ & 1 \\
         & Erdos Renyi 10000 & GPU & $\mathbf{2977.03\pm23.42}$ & 1 \\
         & Random 100 & GPU & $\underline{14.87\pm1.18}$ & 2 \\
         & Random 1000 & GPU & $\underline{116.02\pm1.72}$ & 2 \\
         & Random 10000 & GPU & $\underline{861.53\pm12.72}$ & 2 \\
         & FB15k & GPU & $\underline{3,327.49\pm69.25}$ & 2 \\
         & Wikidata-5m & GPU & $\mathbf{6,559.90\pm128.17}$ & 1\\
        \bottomrule
    \end{tabular}
\end{table}

We observe large runtime variations among different libraries and datasets. However, direct runtime comparisons are challenging because each library generates a distinct set of walks, limiting comparability. Specifically, PyRDF2vec and sparkkgml employ \gls*{bfs} for walk extraction, particularly on smaller graphs, meaning they do not strictly enforce a fixed ``number of walks per entity'' parameter. Instead, the number of walks extracted--and consequently embedded--is naturally capped by the actual number of existing walks available from each entity in the graph. This inherently restricts the total walks embedded during the Word2Vec training process.

The \textit{pyRDF2Vec} implementation, which performs only breadth‐first expansions, exhibits consistently low runtimes on all synthetic graph benchmarks.  On the Barabási–Albert graphs, execution increases modestly from $\sim 2.8$ s at 100 nodes to $\sim 19.5$ s at 10,000 nodes, reflecting its fixed‐size frontier and limited walk count.  Erd\H{o}s–Rényi and uniform random graphs follow a similar trend, with runtimes remaining in the single‐ to low‐tens of seconds even as graph size grows by two orders of magnitude.  The small number of BFS‐only walks explains why \texttt{pyRDF2vec} outperforms full‐random‐walk implementations on these moderate‐scale graphs, but at the cost of reduced walk diversity.

In our experiments, we have seen that sparkkgml does not scale on large graphs. Spark has a lot of different parameters that overall influences the overall query plan execution. However, as the library should nevertheless support multi-node scalability in theory, we have seen that in certain steps during the spark execution, the partition Spark operates on gets reduced to only one partition. The issue arises with the dataframe used for the Word2Vec training. While the number of partitions is set during the method call of to 80, only one partition is used within the training of Word2Vec. A potential reason for this could be the method \textit{struct\_to\_list} which registers a Spark UDF which implements the conversion of a structure datatype to a list.  This overall slows the execution of Spark down and is showcased in the Spark timeout exceptions we receive on our side based on the long running Spark task.

The Java‐based \texttt{jRDF2vec} package, performing full random walks, demonstrates remarkable speed on low‐degree graphs.  On Barabási–Albert networks, runtimes remain nearly constant (\(\sim\!12\) s) across 100–10 000 nodes, since average degree does not grow with network size.  In contrast, on Erd\H{o}s–Rényi graphs—where average degree scales linearly with node count—execution time rises steeply (from 162s to 522s when steeping from 100 to 10,000 nodes).  Uniform random graphs show a moderate increase (11.9s to 21.5s).  These patterns show that \texttt{jRDF2vec} is the fastest way to obtain truly random walks on moderate‐scale, sparse graphs, but its performance degrades on denser graphs.

The GPU‐accelerated \textit{gpuRDF2vec} pipeline is the only implementation that completes full random‐walk extraction on the wikidata-5m knowledge graph ($6,559$ s), as well as all synthetic benchmarks.  For small Barabási–Albert graphs (100 nodes), GPU overhead (data‐transfer, kernel launches) results in slower runtimes (25s) compared to \textit{jRDF2vec}.  However, as graph size and density grow, \textit{gpuRDF2vec} scales smoothly (from 159s to 1,964s when stepping from 1,000 to 10,000 nodes), whereas CPU and Spark implementations either slow dramatically or fail.  This continuous runtime curve, together with its unique ability to process real‐world KBs, positions \textit{gpuRDF2vec} as the clear ``scale champion,'' with further optimization set to close the per‐walk performance gap.

\begin{figure}[t]
    \centering
    \includegraphics[width=0.95\linewidth]{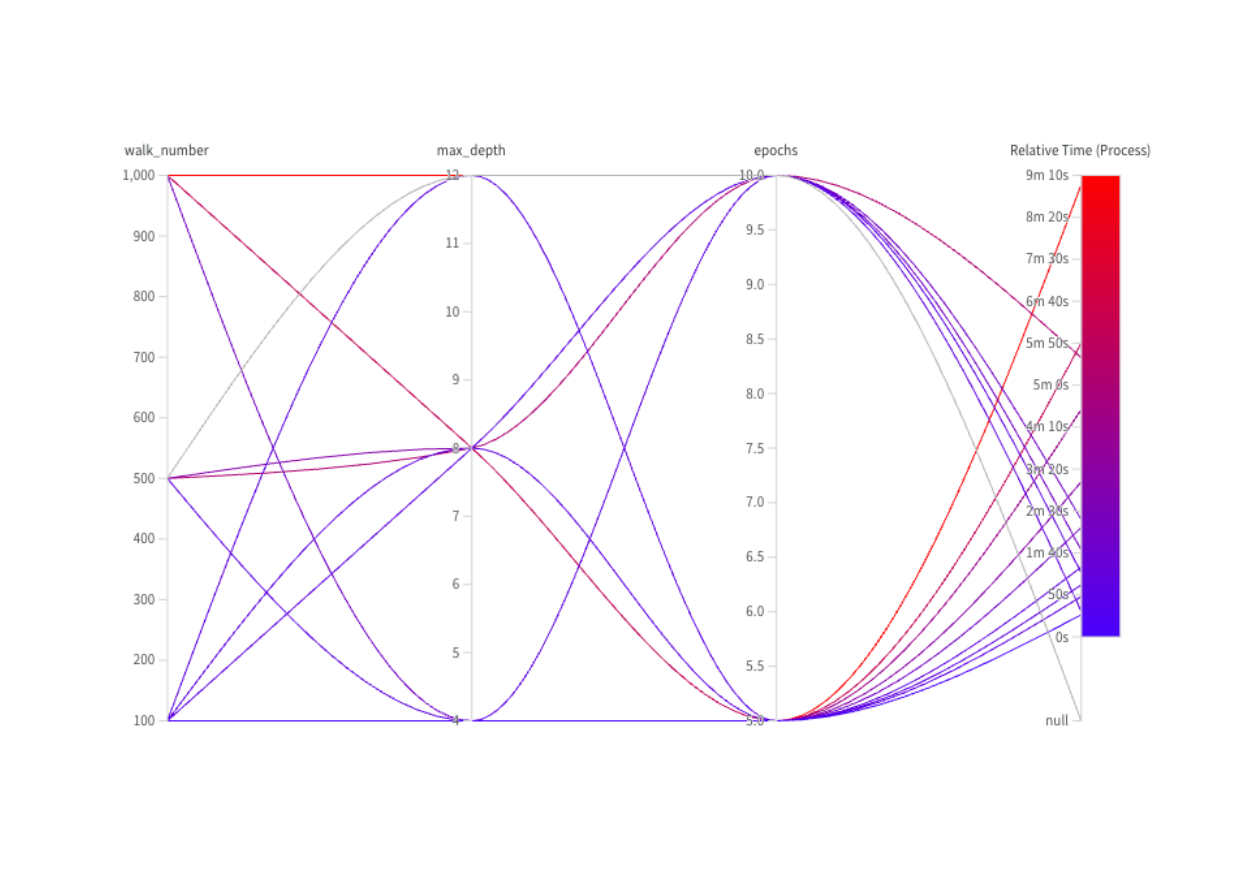}
    \caption{Parallel coordinate of hyperparameter sweep of gpuRDF2vec runs on Erd\H{o}s Rényi 1000 graph}
    \label{fig:gpuRDF2vec_hpt}
\end{figure}

\autoref{fig:gpuRDF2vec_hpt} illustrates that the total processing time of our GPU‐accelerated RDF2Vec implementation is dominated by the number of random walks and the number of training epochs, both of which exhibit an approximately linear relationship with runtime. Specifically, raising \texttt{walk\_number} from 100 to 1000 increases the wall‐clock time by almost an order of magnitude, and doubling \texttt{epochs} from 5 to 10 yields a near doubling of total execution time. By contrast, the choice of \texttt{max\_depth} has only a secondary, non‐monotonic effect: walks limited to a medium depth ($8$) achieve slightly lower runtimes than either very shallow $(4)$ or very deep $(12)$ walks  which is rather remarkable because that latter parameter has been shown to have the largest impact on the qualitative results obtained with RDF2vec. This suggests that overly shallow traversals underutilize the GPU’s parallelism, while excessively deep walks incur additional per‐step overhead. In practice, then, tuning \texttt{walk\_number} and \texttt{epochs} is the most effective way to control overall runtime, whereas setting \texttt{max\_depth} to an intermediate value offers the best balance between embedding quality and computational cost.

\section{Limitation and Future Work} \label{cha:LimitationAndFutureWork}
While our GPU-accelerated RDF2Vec pipeline demonstrates significant performance gains, several limitations remain and motivate future extensions:
 
Our current implementation relies on the standard (unordered) Word2Vec algorithm; it does not yet support order-aware variants (e.g.\ Structured Skip-Gram \cite{Ling2015,Portisch2021}) or binary (hash) embeddings~\cite{faria2024towards}. However, thanks to the modular Python-based design, we can integrate them by adding an order-aware or binary embedding module in place of the existing Word2Vec routines.

At present, we parse non-NT RDF serializations (e.g.\ Turtle, RDF/XML) via RDFlib, which loads the entire graph into CPU memory and thus becomes a bottleneck for large datasets. For future versions, we plan to develop a streaming, GPU-friendly RDF parser—perhaps leveraging RAPIDS cuIO—to ingest diverse RDF formats without sacrificing scalability. Although we stream cuDF partitions into PyTorch tensors, we currently assume all partitions fit in the available VRAM simultaneously. A more robust strategy would dynamically spill oversized partitions to host memory or temporary files, and prefetch or evict them based on utilization, further reducing OOM risk and manual tuning.

Certain operations, such as context window indices or negative-sampling tables are, statically precomputed at startup. Making these computations adaptive (e.g.\ recomputing windows for changing graph statistics or sampling budgets) would improve robustness to heterogeneous datasets. While our GPU kernels accelerate most of the training loop, Python overhead in the model architecture remains non-negligible for very large models. We will continue profiling and migrate critical sections to custom CUDA/C++ extensions where appropriate.

To facilitate reproducibility and hyperparameter sweeps, we intend to integrate with logging platforms such as MLflow or Weights \& Biases. Automatic metric logging, artifact storage, and configuration tracking will streamline large-scale evaluations.  Finally, we aim to extend our framework beyond Word2Vec to train RDF2Vec embeddings using transformer-based language models (e.g.\ RoBERTa or GPT variants) on GPU.~\cite{agozzino2021sesame}
This extension might be especially interesting for large graph corpora where many tokens are present. This potential extension will allow us to compare classical and contextualized embeddings within a unified, scalable infrastructure.

Biased and weighted random walks have been shown to improve embedding quality by prioritizing semantically relevant paths, thereby capturing more meaningful graph structures in the resulting vector representations \cite{cochez2017biased,mukherjee2019graphnodeembeddingsusing,Boeckling2025}. The planned extension will introduce support for configurable edge weights and custom biasing strategies, enabling users to tailor walk generation to specific knowledge graph characteristics or downstream tasks. To ensure scalability on large knowledge graphs, we will explore graph flooding techniques for weight propagation, allowing for local weight computation besides centrality measures.



\section{Conclusion}
Of all the RDF2Vec implementations evaluated, \textbf{only} \texttt{gpuRDF2vec} was able to complete the full random‐walk extraction on the real‐world FB15k knowledge graph.  Both CPU‐based random‐walk approaches (e.g.\ \texttt{jRDF2vec}) and the BFS‐only packages (\texttt{pyRDF2vec} and \texttt{sparkkgml}) failed to finish within reasonable time or resource limits.  This result alone elevates \texttt{gpuRDF2vec} to the role of \emph{scale champion}: it is the sole system in our study capable of processing a graph with on the order of $10^5$ entities and $10^6$ edges end‐to‐end. Moreover, the implementation does scale well for emph{longer} walks, which is a crucial advantage since the walk length has been identified as a crucial parameter for the embedding quality in the past.

Although the raw \emph{seconds per walk} metric still favors a tuned CPU implementation (\texttt{jRDF2vec}) on small to moderate synthetic graphs (up to $10^4$ nodes), the GPU approach exhibits the only \emph{smooth, monotonic growth} in runtime as graph size and density increase.  In contrast, CPU and Spark runs either blow out in execution time or terminate prematurely.  Therefore, the GPU method’s throughput — is the only one that remains strictly positive and well‐behaved at large scale. Looking forward, the true promise of \texttt{gpuRDF2vec} lies in its \emph{optimization headroom}.  By fusing kernels, overlapping data transfers, and adopting zero‐copy memory techniques, we can dramatically reduce the per‐walk overhead. Such improvements will not only close the gap on small graphs but will also further widen \texttt{gpuRDF2vec}’s advantage on massive knowledge graphs.

In summary, while CPU‐based RDF2vec remains attractive for quick prototyping on small graphs, and BFS‐only extraction offers a compelling trade‐off for extremely fast but less diverse walks, \texttt{gpuRDF2vec} stands alone as the only approach proven to ``scale'' to larger knowledge graph sizes.  This makes it the clear foundation for future work on truly large‐scale, high‐throughput RDF2vec embedding.

\section*{Resource Availability Statement}
We provide a detailed overview of the overall runtime comparison of our experiment in the \href{https://wandb.ai/m_boeck/rdf2vec_runtime_comparison}{WANDB project}, model comparisons, hardware utilization, and reproducibility artifacts. Our code under the following \href{https://github.com/MartinBoeckling/rdf2vecgpu}{Github repository} is published under the MIT license. We are committed to provide 
a Python package on pypi together with installation instructions to run successfully on CUDA 12.x devices.

%
%
%
\newpage
\bibliographystyle{splncs04}
\bibliography{mybibliography}
%




\end{document}